\documentclass[10pt,twocolumn,letterpaper]{article}

\usepackage[pagenumbers]{wacv} 
\usepackage[accsupp]{axessibility} 
\usepackage{amsmath}
\usepackage{amssymb}
\usepackage{booktabs}
\usepackage{circledsteps}
\usepackage{tabularray}
\usepackage{rotating}
\usepackage{pifont}
\usepackage{xcolor}
\usepackage{graphicx}
\usepackage{color, colortbl}
\definecolor{lightgray}{gray}{0.8}
\usepackage{float}
\usepackage{algorithm}
\usepackage[noend]{algpseudocode}
\usepackage{multirow}
\usepackage[accsupp]{axessibility} 
\usepackage{subfiles}
\usepackage{makecell}
\usepackage{threeparttable}
\usepackage[pagebackref,breaklinks,colorlinks]{hyperref}

\newcommand{\fer}{FER\xspace}
\newcommand{\hpr}{$F_{hpr}$\xspace}
\newcommand{\chpr}{$C_{hpr}$\xspace}
\newcommand{\lpr}{$F_{lpr}$\xspace}
\newcommand{\clpr}{$C_{lpr}$\xspace}
\newcommand{\fc}{$F_{fc}$\xspace}
\newcommand{\cfc}{$C_{fc}$\xspace}
\newcommand{\vpl}{$V_{pl}$\xspace}
\newcommand{\fu}{$F_{u}$\xspace}
\newcommand{\cmark}{\textcolor{blue}{\ding{51}}} 
\newcommand{\xmark}{\textcolor{red}{\ding{55}}}

\usepackage[capitalize]{cleveref}
\crefname{section}{Sec.}{Secs.}
\Crefname{section}{Section}{Sections}
\Crefname{table}{Table}{Tables}
\crefname{table}{Tab.}{Tabs.}

\begin{document}

\title{Facial Expression Recognition with Controlled Privacy Preservation and \\ Feature Compensation}

\author{
Feng Xu$^{1,2}$, 
David Ahmedt-Aristizabal$^{2}$
Lars Petersson $^{2}$
Dadong Wang $^{2}$
Xun Li$^{2}$ \\
$^{1}$UNSW Sydney,
$^{2}$CSIRO Data61, Australia\\ 
{\tt\small \{feng.xu,~david.ahmedtaristizabal,~xun.li\}@data61.csiro.au,}
}

\maketitle

\begin{abstract}

Facial expression recognition (FER) systems raise significant privacy concerns due to the potential exposure of sensitive identity information. 
This paper presents a study on removing identity information while preserving FER capabilities.
Drawing on the observation that low-frequency components predominantly contain identity information and high-frequency components capture expression, 
we propose a novel two-stream framework that applies privacy enhancement to each component separately. 
We introduce a controlled privacy enhancement mechanism to optimize performance and a feature compensator to enhance task-relevant features without compromising privacy. 
Furthermore, we propose a novel privacy-utility trade-off, providing a quantifiable measure of privacy preservation efficacy in closed-set FER tasks.
Extensive experiments on the benchmark CREMA-D dataset demonstrate that our framework achieves 78.84\% recognition accuracy with a privacy (facial identity) leakage ratio of only 2.01\%,  highlighting its potential for secure and reliable video-based FER applications.
We encourage the readers to visit the project page: \url{https://fengxxu.github.io/ppfer/}.
\end{abstract}


\section{Introduction}
\label{sec:intro}

Facial Expression Recognition (\fer) has been attracting growing attention due to its great value in various fields, such as medical diagnosis~\cite{ahmedt2018deep,sonkusare2019detecting,yang2021undisturbed,cantrill2024orientation}, human-robot interaction~\cite{liu2017facial,faria2017affective, serfaty2023generative} and driver fatigue detection~\cite{albu2008computer,du2020vision}. 
Protecting facial identity in these sensitive application areas is crucial, not only because it is required by law~\cite{act1996health,voigt2017eu}, but also as an ethical and moral obligation to foster user trust and encourage participation. 
AI has revolutionized numerous industries with its transformative capabilities. However, it has yet to be widely adopted in human-related fields, primarily due to significant privacy concerns. As AI algorithms in these fields depend on extensive human data for training and optimization, securing sensitive data and limiting access to authorized personnel only is essential.

\begin{figure}[!t]
   \centering
   \includegraphics[width=1.0\linewidth]{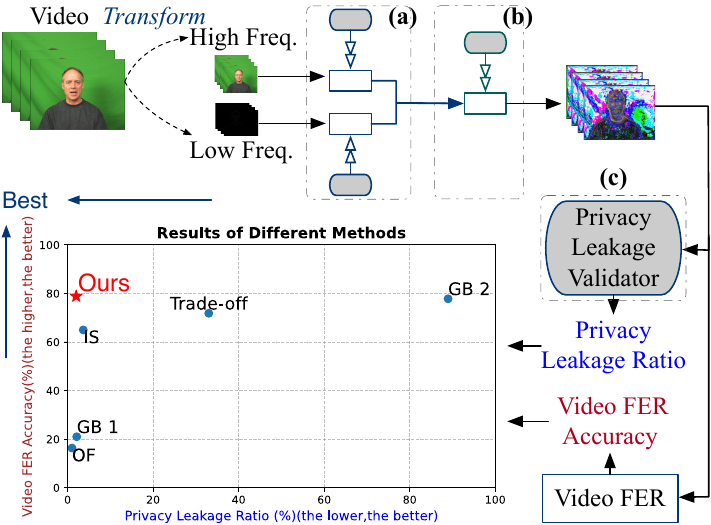}
   \caption{Our framework: 
   \textbf{(a)}. Controlled privacy-preservation in the high- and low-frequencies (Freq.) with \textbf{(b)}. Controlled feature compensation and \textbf{(c)}. Privacy leakage validation.
   The results show the performance of different privacy preservation approaches with video-based \fer on privacy leakage ratio and video \fer accuracy. 
   GB 1 \& 2, OF, Trade-off and IS are short for Gaussian Blurring 1 \& 2, Optical Flow, Trade-off framework and Image Swapping, respectively.
   }
   \label{fig:intro}
\vspace{-12pt}
\end{figure}

Researchers have extensively studied privacy preservation for various utility tasks, from full-body behavior recognition to facial analysis.
In this study, we focus on the latter task, where visual obfuscation techniques have been used to protect privacy. The preservation of facial privacy can be categorized into two scenarios~\cite{osorio2021stable}: {\it closed-set}, where the set of all possible identities in the observed data is known to the system, and {\it open-set}, where some identities in the observed data are not previously known to the system. Our work specifically addresses the closed-set scenario.
We summarize existing approaches into four categories for both scenarios based on different methodologies employed.
The first category of approaches~\cite{wu2020privacy,yan2020low,wang2022privacy,ji2022privacy} distorts source data using techniques like blurring, pixelation, or noise addition to protect privacy, but these methods also disrupt facial features, compromising tasks such as facial biometrics recognition and \fer.
The second category focuses on face reconstruction~\cite{mai2018reconstruction}. While these two categories can partially preserve privacy, they remain vulnerable to recovery attacks~\cite{razzhigaev2020black}.
The third category synthesizes new images, including face generation~\cite{chen2018vgan, ashwin2023preserving} and face swapping~\cite{xu2022mobilefaceswap}, to preserve expressions or replace faces. Nevertheless, existing works on static images \cite{lopez2022deepfakes} highlight that facial expression features are not always transferred effectively.
The last category is not explicitly designed for privacy protection but achieves it as a byproduct. 
For example, video data can be converted to optical flow \cite{allaert2022comparative}, or infrared thermal images can be used to classify expressions in low-light settings~\cite{bhattacharyya2021deep}.
While optical flow effectively protects privacy, it complicates downstream tasks. The infrared thermal approach performs well in low light but struggles with dataset domain shifts~\cite{prasad2023mobilenetv3, chatterjee2024moth}.
Even though these works are not explicit about either open-set or closed-set scenarios, we observe a lack of mechanisms for evaluating privacy preservation for specific attributes, making quantitative comparisons difficult. 

\fer from video data presents unique challenges in balancing expression classification accuracy with privacy preservation. It requires removing facial identity features while retaining facial expression details for accurate recognition. Traditional methods~\cite{dave2022spact, shi2023videoflow,xu2022mobilefaceswap} often struggle to effectively remove personal identity features without compromising facial expression dynamics,  leaving privacy vulnerable to manual review. Moreover, utility task accuracy typically declines post-preservation. Although privacy enhancement is necessary, it inevitably degrades facial expression features, necessitating feature enhancement approaches to maintain performance. Despite prior efforts, there remains a lack of independent mechanisms to quantitatively assess privacy leakage.

In response to these challenges, our proposed framework, as shown in Figure~\ref{fig:intro}, exploits the wavelet transform to remove personal identity features from the low-frequency and high-frequency components and separately removes the privacy identity features in low- and high-frequencies. The transformation is inspired by \cite{gao2011comparison}, which illustrates that the facial expression frequency band is higher than the facial identity frequency band. This has also been proven by subsequent works~\cite{kumar2011emotion, bate2015independence}. In video clips, such as those depicting a transition from a neutral to a happy facial expression and back, personal identity features persist consistently across frames as low-frequency information, while the face expression dynamics, which constitute high-frequency information, will vary across frames. The frequency distinction, where facial expression information is carried in high frequency and identity information is mostly presented in low frequency, suggests a potential method to preserve expression while removing identity.
Our framework adopts a dual approach, removing facial identity from both low- and high-frequency components, as facial identity features may not reside exclusively in the low-frequency range. To optimize this, we introduce privacy enhancement controllers that adjust the privacy enhancer for each frequency by maximizing identity recognition loss.
After privacy enhancement, the two frequency bands are fused, and frames are reconstructed via inverse transformation. This is followed by a feature compensator,  inspired by~\cite{li2024color}, which enriches non-identity features, enhancing the fidelity of facial expression.
In addition, we introduce a new mechanism that quantitatively assesses privacy preservation at the final step for performance validation.
Concurrently, the utility task of \fer has been implemented as a classifier at the final step operating on the reconstructed videos.
The contributions of our paper are three-fold:
\begin{itemize}
\vspace{-3pt}
    \item We propose a new framework for privacy-preserved video-based facial expression recognition by using a dual approach, which separately removes existing identity privacy from high- and low-frequency components. This process is further enhanced by two dedicated privacy preservation controllers.
    \vspace{-3pt}
    \item We introduce a feature compensator that compensates the non-identity features to enhance the performance of the utility task (video-based \fer in our case), ensuring robust performance even after privacy enhancement.
    \vspace{-3pt}
    \item We propose a novel mechanism that evaluates the effectiveness of privacy preservation by comparing the ground truth identity labels with the output produced from a facial identity classification network after removing identity information in the closed-set scenario. We believe that this mechanism can be extended to other privacy attributes for specific validation.
\end{itemize}
\vspace{-3pt}

\section{Related Work}\label{rw}
\vspace{-5pt}
\noindent\textbf{Deidentification.}
To conceal the visual personal identity, PartialFace~\cite{mi2023privacy} hides visual information by pruning low-frequency components and applying random frequency components during training and inference to impede recovery. Building on PartialFace, MinusFace~\cite{mi2024privacy} further subtracts the features from original faces and their regenerated versions to produce protective face representations. While these methods effectively preserve privacy within the closed-set scenario, they are limited in face recognition tasks. 
Yoon et al.~\cite{yoon2012altered} protect privacy by altering biometric information to prevent recognition of biometric features. Techniques proposed by~\cite{4587369,brkic2017know} focus on de-identification, removing or replacing personal identifiers in biometric data. Additionally, Orekondy et al.~\cite{orekondy2018connecting} remove privacy-sensitive features from biometric images and construct visual redactions. 
Osorio-Roig et al.~\cite{osorio2021stable} evaluate biometric privacy preservation approaches in both open-set and closed-set scenarios. These methods utilize various approaches to preserving privacy, but they fail to validate their privacy-enhanced images to demonstrate that the identity cannot be accurately recognized by computer vision models

\noindent\textbf{Privacy-enhancement of Facial Expression}
A weighted supervised contrastive learning approach is used for privacy-enhancement in recognizing facial expression embedding from images~\cite{rosberg2021comparing}. Adversarial learning techniques are utilized by~\cite{narula2020adversarial,narula2020preserving} to train Convolutional Neural Networks (CNNs), preserving facial identity in images while ensuring high \fer accuracy. 
Leibl et al.~\cite{leibl2023identifying} subtract and blend facial identity from images, swapping faces and retaining the associated facial expression feature. 
Similarly, image synthesis is used to enhance privacy while preserving \fer performance, as demonstrated in \cite{rahulamathavan2015efficient,chen2018vgan}. 
IRFacExNet~\cite{bhattacharyya2021deep} recognizes facial expressions from infrared thermal images, using a cosine annealing learning scheduler~\cite{loshchilov2016sgdr} to enhance overall performance. These works implement privacy-preserved \fer based on images.

\noindent\textbf{Privacy-enhancement of Soft-biometrics}
Wu et al.~\cite{wu2020privacy} propose an adversarial training framework to learn a privacy enhancement function that protects identity, gender, ethnicity, and other soft biometrics in closed-set scenarios. 
SPAct~\cite{dave2022spact} builds upon this work by introducing a self-supervised privacy branch that addresses additional human-related privacy attributes in open-set scenario.
Furthermore, TeD-SPAD~\cite{fioresi2023ted} extends the SPAct framework to anomaly detection. These studies~\cite{narula2020preserving, narula2020adversarial, low2022adverfacial, wu2020privacy, dave2022spact, fioresi2023ted} focus on trade-off frameworks that aim to balance privacy preservation with utility tasks, providing optimal solutions that reconcile these competing objectives.

\noindent\textbf{Recognition of Facial Expression from Videos.}
Video-based \fer has been extensively investigated, with recent 
works addressing challenges such as changes unrelated to facial expressions and overfitting in recognition models~\cite{li2020deep,adyapady2023comprehensive}.
Many studies~\cite{fan2016video,kossaifi2020factorized} utilize 3D CNNs to extract features from videos, while others, such as EC-STFL~\cite{jiang2020dfew} combine Recurrent Neural Networks with CNNs for dynamic facial expression recognition.
Advanced techniques, including intensity-aware loss with the global convolution-attention block\cite{li2023intensity}, have been developed to further enhance recognition performance. 
M3DFEL~\cite{wang2023rethinking} addresses short-term and long-term temporal relationships by generating 3D instances using 3D CNNs for feature extraction and dynamically aggregating long-term instances.

\noindent\textbf{Feature Compensation.}
DAFE~\cite{chen2019improving} enhances utility task-related features by reusing low-level image features derived from spatial details and pixel-wise features, leveraging channel-wise long-range relationships. SAFECount~\cite{you2023few} employs location regression on the original images and integrates this with a similarity map to improve accuracy, guided by a similarity metric. 
COSE~\cite{li2024color} enhances images by learning from color shifts through over- and under-exposure of the original images. 
U-ViT~\cite{bao2023all} treats input patches as tokens, incorporating temporal, conditional, and noisy images to generate new outputs while utilizing long skip connections within its framework, significantly enhancing the performance of image diffusion models.

\section{Method}
\label{method}

\begin{figure*}[ht]
    \centering
    \includegraphics[width=0.99\textwidth]{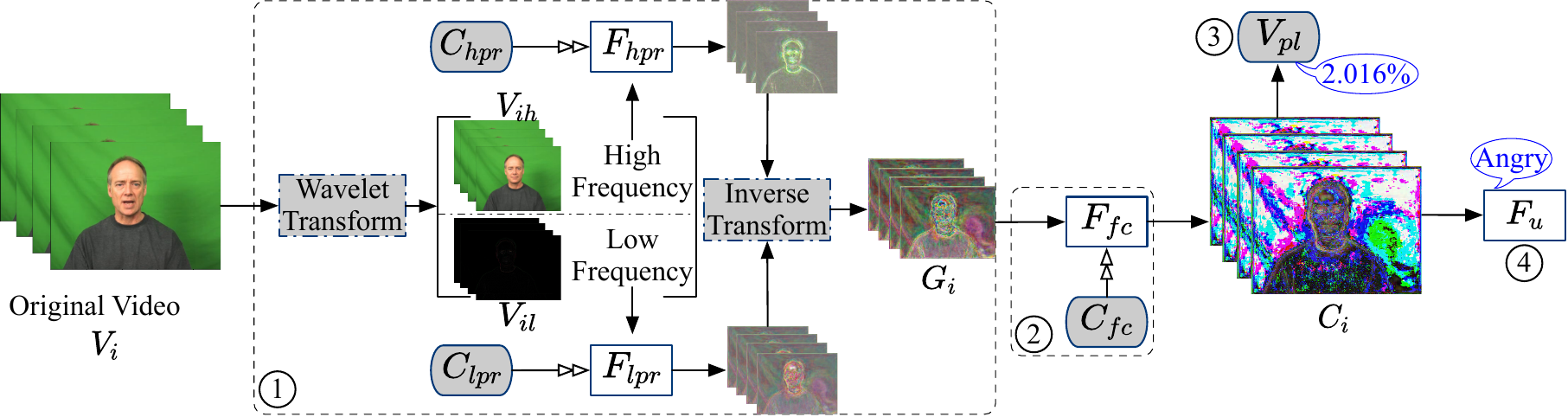}
    \caption{Illustration of our framework. \Circled{1} Controlled privacy-preservation: the original video ($V_i$) is transformed into high- and low-frequency components ($V_{ih}$ and $V_{il}$), followed by privacy enhancement, which includes a privacy enhancer (\hpr and \lpr) and its controller to strengthen privacy enhancement (\chpr and \clpr) for low- and high-frequency components respectively. The privacy-preserved high- and low-frequency frames are then combined and inverse-transformed into new video frames ($G_i$) for the next step. \Circled{2} Controlled feature compensation involves a feature compensator (\fc) and its controller (\cfc), which regulates the compensation of specific features. \Circled{3} Privacy leakage validation uses a privacy leakage validator (\vpl) to determine the proportion of identity features that can be recognized to measure the performance of privacy preservation. \Circled{4} Utility task (\fu) learns from the feature-rich video frames ($C_i$))on its main task, such as FER in our case. All grey components are frozen during training, and all controllers are not involved in inference.}
    \label{fig:framework}
    \vspace{-6pt}
\end{figure*}

In previous literature~\cite{dave2022spact,shi2023videoflow,xu2022mobilefaceswap}, we observe that most researchers treat privacy preservation and main utility tasks as a tangled problem, where their methodologies struggle to maintain the high performance of both tasks. In this research, we address such a dilemma by decoupling the two tasks in a way that two separate networks/mechanisms, namely the privacy preservation controller and feature compensator, have been designed and employed at the two tasks, respectively, thus achieving good performance for both.

\subsection{Model Overview}

The proposed framework consists of four key components, as illustrated in Figure~\ref{fig:framework}: 
\Circled{1} controlled privacy-preservation, 
\Circled{2} controlled feature compensation, 
\Circled{3} privacy leakage validation (\vpl), and 
\Circled{4} the utility task (\fu), focusing on video-based \fer. 
In the framework, \hpr and \chpr refer to the high-frequency privacy enhancer and its identity budget controller, respectively. Similarly, \lpr and \clpr denote the low-frequency privacy enhancer and its identity budget controller. \fc and \cfc represent the non-identity feature compensator and its controller.
The pre-trained controllers (\chpr, \clpr and \cfc) and the privacy leakage validator (\vpl) are frozen during each training process to maintain their integrity and effectiveness.

\subsection{Controlled Privacy-Preservation}
\begin{algorithm}[ht]
\caption{Controlled privacy preservation training}\label{alg:cpp}
    \begin{algorithmic}[1]
        \Statex \textbf{Initialization:} \chpr, \clpr, \hpr and \lpr load pre-trained weights
        \Statex \textbf{Input:} A batch of training videos, facial expression labels and identity labels $(V_i,E_i,I_i)\in D$, where $D$ is the training dataset with $n$ batch of videos and $i \in \{1,...,n\}$
        
        \For{$epoch = 1$ to $N_1$} 
            \State Shuffle the training data
            \For{$i=1$ to $n$}
                \State \hspace*{-1mm}Wavelet transform $V_i$ and get high-($V_{ih}$) and low-($V_{il}$) freq.;
                \State \hspace*{-1mm}\chpr and \clpr set as .eval(), \hpr and \lpr set as .train();
                \State \hspace*{-1mm}\textit{\footnotesize \# Forward pass:}
                \State \hspace*{-1mm}High freq. output $y_{ih}: = C_{hpr}(F_{hpr}(V_{ih}))$;
                \State \hspace*{-1mm}Low freq. output $y_{il}: = C_{lpr}(F_{lpr}(V_{il}))$;
                \State \hspace*{-1mm}\textit{\footnotesize \# Backward pass:}
                \State \hspace*{-1mm}Update parameters of \hpr and \lpr via maximizing the cross-entropy losses of $(y_{ih},I_i )$ and $(y_{il},I_i)$;
            \EndFor
        \EndFor
        
        \For{$i=1$ to $n$} \textit{\footnotesize \# Remove training set identity}
            \State \hspace*{-1mm}$G_i:=$ inverse transform $F_{hpr}(V_{ih})$ and $F_{lpr}(V_{il})$; 
        \EndFor
    \Statex \textbf{Output:} The privacy-preserved training videos $G_i$, $(G_i,E_i,I_i) \in D_G$, where $D_G$ is the privacy-preserved training dataset.
    \end{algorithmic}
\end{algorithm}

Previously,  we established that while a person's identity information persists across all frames in a video, facial expression undergoes dynamic changes.  
Thus, dividing the video into high- and low-frequency components allows for differential treatment of identity and expressive information. High-frequency components contain facial expressive information, whereas low-frequency components retain identity-related features. 
Because the identity-related features contained in the two are different, the removal of privacy at different frequencies should be different.
Consequently, at the beginning of the framework, the original video is transformed into high- and low-frequency components ($V_{ih}$ and $V_{il}$) to preserve privacy separately.

In our framework, controlled privacy-preservation, as illustrated in Algorithm~\ref{alg:cpp}, involves two main functions: an identity budget controller (\chpr or \clpr) and a privacy enhancer (\hpr or \lpr). 
The identity budget controller is implemented as a CNN-based classification model that is pre-trained on identity label datasets using cross-entropy loss. It classifies the identity of the person appearing in each video frame. On the other hand, the privacy enhancer, an encoder-decoder model, is pre-trained with $L1$ reconstruction loss similar to~\cite{dave2022spact}.
In our framework, the initial parameters of \chpr and \clpr are identical, as are those of \hpr and \lpr. 
During training, \chpr and \clpr are frozen, and for each frequency, the transformed video serves as input for the privacy enhancer. The output of the privacy enhancer is then used as input for the identity budget controller, as shown in lines 7 and 8 of Algorithm~\ref{alg:cpp}. We compute the cross-entropy loss between the output of the identity budget controller and the identity labels, optimizing only the parameters of the privacy enhancer.
The outputs from the high- and low-frequency privacy-preservation processes are then integrated and inverse transformed to reconstruct a video, $G_i$ in Algorithm~\ref{alg:cpp}, which serves as the input for the subsequent components within the framework.

\subsection{Controlled Feature Compensation}

\begin{algorithm}[!t]
\caption{Controlled feature compensation training}\label{alg:cfc}
\begin{algorithmic}[1]
\Statex \textbf{Initialization:} \fc and \cfc load pre-trained weights
\Statex \textbf{Input:} $(G_i,E_i,I_i) \in D_G$ from Algorithm~\ref{alg:cpp}
\For{$epoch = 1$ to $N_2$}
    \State Shuffle the training data
    \For{$i=1$ to $n$}
    \State \hspace*{-1mm}\cfc sets as .eval(), \fc sets as .train();
    \State \hspace*{-1mm}\textit{\footnotesize \# Forward pass:}
    \State \hspace*{-1mm}$G_{ci}:=F_{fc}(G_i)$;
    \State \hspace*{-1mm}$e_{gci}:= C_{fc}(G_{ci})$;
    \State \hspace*{-1mm}\textit{\footnotesize \# Backward pass:}
    \State \hspace*{-1mm}Update parameters of \fc via minimizing the cross-entropy loss $(e_{gci}, E_i)$
    \EndFor
\EndFor
\For{$i=1$ to $n$} \textit{\footnotesize \# Compensate training set features}
    \State \hspace*{-1mm}$C_i:=F_{fc}(G_i)$;
\EndFor
\Statex \textbf{Output:} Feature compensated training set with facial expression labels $(C_i,E_i) \in D_G$
\end{algorithmic}
\end{algorithm}

Algorithm~\ref{alg:cfc} illustrates the controlled feature compensation training approach.
The feature compensation module in our framework consists of two components: \fc and \cfc. \fc uses the U-ViT model~\cite{bao2023all}, which integrates Vision Transformer (ViT) as the backbone for image generation with the diffusion model. 
This component enhances the utility-task related features for the privacy-preserved video frames. It intends to operate frame-by-frame to generate enhanced visual representations for downstream utility tasks while suppressing specific identity-related features. \cfc is a classification model employing ResNet50, pre-trained on a facial expression dataset collected in real-world conditions, using cross-entropy loss. 
This component learns from a wide variety of data collected in real-world conditions, thus ensuring that the guided feature compensation enhances the common facial expression features without being biased by specific identity-related features.

During the training process of \fc, the pre-trained \cfc is frozen and provides feedback for \fc. 
Optimization involves training \fc using cross-entropy loss to guide the generation of facial expression feature-rich frames. This approach ensures that the feature compensation enhances the expressive content of privacy-preserved videos and generates feature-compensated video, $C_i$ in Algorithm~\ref{alg:cfc}, while not affecting the privacy enhancement quality in the video.
 
\vspace{-3pt}
\subsection{Privacy Leakage Validation}
\vspace{-3pt}
As previously noted, we create the leakage validation mechanism based on identity recognition accuracy using a pre-trained classification network that registers all facial identities.
Our privacy leakage validator \vpl uses ResNet50, pre-trained on the same dataset with identity labels as \chpr and \clpr, to recognize a random frame from the feature-compensated video.
By maintaining consistency between \chpr, \clpr, and \vpl, we ensure that the privacy attribute being validated for leakage is the same one that was previously preserved.
The \vpl measures privacy leakage by assessing the accuracy of recognizing facial identity features from privacy-preserved videos. A higher accuracy of \vpl indicates greater residual identity information in the video, thus higher privacy leakage.
Consequently, the \vpl transfers its identity recognition accuracy into a measure of privacy leakage, providing a quantitative mechanism for comparing the performance of various privacy preservation approaches.

The privacy leakage validation process requires explicit privacy categories, such as identity, skin color, and gender. This enables an assessment of how effectively the privacy preservation method addresses specific privacy categories, regardless of the utility task.

\subsection{Facial Expression Recognition}
R3D~\cite{tran2018closer} is selected as the network for the video-based \fer. Previous works~\cite{goncalves2023versatile, ryumina2022search} have demonstrated good performance in recognizing the expressions on our selected dataset. 
In our framework, the downstream task, \fu, does not affect the upstream tasks, including privacy preservation and feature compensation. 
Training \fu on $V_i$ and $C_i$ can demonstrate how our privacy preservation affects video-based FER by comparing the recognition accuracy of $F_u(V_i)$ and $F_u(C_i)$.

\section{Experiments}

\vspace{-3pt}
\subsection{Experimental Setup}\label{exp_setup}
\vspace{-3pt}
Our framework operates in a closed-set scenario, requiring datasets that provide ground truth facial expression labels for classification and facial identity information for validating privacy protection.
We selected the CREMA-D~\cite{cao2014crema}, which includes 7,442 video clips from 91 actors displaying six categories of facial expressions. For our experiments, we used 2,232 clips for testing (CREMA-D testing set) and the remaining clips for training (CREMA-D training set). 
The identity budget controllers and privacy leakage validator are pre-trained on the CREMA-D training set.
The non-identity feature controller was pre-trained on 3,600 clips from the DFEW dataset~\cite{jiang2020dfew}, which contains 16,372 in-the-wild video clips from movies featuring 7 categories of facial expressions.

\subsection{Implementation} 
As illustrated in Figure~\ref{fig:framework}, our framework employs the wavelet transform to decompose the original video into high-frequency and low-frequency components. The assumption is that identity information is primarily contained in the low-frequency part, as the same individuals perform facial expressions throughout the video. Conversely, facial expressions representing dynamic changes are captured in the high-frequency part, except in the case of static images. 
To address privacy enhancement, our framework splits the process into two branches, each consisting of a privacy enhancer and an identity budget controller. Then, we follow a two-step training procedure. First, we train ResNet50 on the CREMA-D training set to recognize the identities of the 91 actors, serving as the identity budget controllers. These controllers achieve 100\% accuracy on the CREMA-D testing set. Then we train the privacy enhancer, during which time its corresponding controller remains fixed throughout the training of privacy enhancer, making sure they effectively remove identify-related features.
Both privacy enhancers use the U-Net~\cite{ronneberger2015u} as the backbone and are pre-trained using $L_1$ reconstruction loss, similar to the anonymization function in SPAct~\cite{dave2022spact}. 
Finally, the wavelet inverse transform is applied to the output of the high- and low-frequency privacy enhancers to reconstruct the video frames.  

In the controlled feature compensation, 
\fc follows the U-ViT architecture~\cite{bao2023all} and uses the ``ImageNet 256x256 (U-ViT-H/2)'' pre-trained weights. 
The \cfc is pre-trained with cross-entropy loss on the DFEW dataset, using one random frame from each video.
During training, \cfc remains frozen while \fc is updated. 

The \vpl utilizes ResNet50 with pre-trained parameters, the same as those used for \chpr and \clpr, for inference. Given that all frames within a feature-compensated video contain the same identity, a randomly selected frame from the video is used as input of \vpl. \vpl requires the ground truth privacy feature labels to validate prediction accuracy. 

The utility task \fu employs the R3D, pre-trained on the Kinetics-400~\cite{kay2017kinetics} dataset and trained with cross-entropy loss, to perform video-based facial expression classification.
The pre-training of \chpr, \clpr, \cfc, and \vpl, as well as the training of \hpr, \lpr,\fc, and \fu, uses the CREMA-D training set or its generated set.

\vspace{-3pt}
\subsection{Baseline Comparison}
\vspace{-3pt}
To evaluate our framework, we first train \fu on the original dataset to check the performance of \fu, named ``No Privacy Preservation'' in Table~\ref{main_result}. Then, we implement several baselines:

\noindent\textbf{Distortion.}
We implement a Gaussian blur approach as the baseline to preserve identity. The formulation is $G(x,y)= \frac{1}{2\pi \sigma^2}e^{ - \frac{x^2+y^2}{2\sigma^2} }$, where $x$ and $y$ are the horizontal and vertical distances from the center, and $\sigma$ is the standard deviation. As $\sigma$ controls the degree of blurring, we implement two baselines aiming to align with our performance of either privacy enhancement or utility task to compare the trade-off.
Firstly, we set the degree of privacy leakage (2.150\%) after Gaussian blur to be closed to our approach (2.016\%), named ``Gaussian Blur 1'' in Table~\ref{main_result}, and we observe the performance of \fu. 
Secondly, we adjust the Gaussian blur, named ``Gaussian Blur 2'' in Table~\ref{main_result}, so that the accuracy (77.778\%) of the utility task reaches the same level as our results (78.843\%), and then we observe the privacy leakage ratio.

\noindent\textbf{Privacy-enhancement via adversarial learning.}
We also implement SPAct~\cite{dave2022spact}, a bi-objective optimization framework, a state-of-the-art self-supervised privacy enhancement approach building upon an adversarial learning framework~\cite{wu2020privacy}, and we name it as ``Trade-off Framework'' in Table~\ref{main_result}.

\noindent\textbf{De-identification.}
The approach involves transforming the image or video data into another modality, such as converting video to optical flow. We select a state-of-the-art framework, VideoFlow~\cite{shi2023videoflow}), as the de-identification baseline, named as ``optical flow'' in Table~\ref{main_result}.

\noindent\textbf{Image swapping.}
This approach aims to demonstrate that face swapping could preserve facial landmarks while hiding personal identity.
We utilize a state-of-the-art approach~\cite{xu2022mobilefaceswap} that utilizes a face not in the dataset to swap all faces appearing in the CREMA-D testing set. \vpl and \fu perform the same tasks as in our framework on the output of face swapping. The results are shown as ``Image Swapping'' in Table~\ref{main_result}.

\vspace{-5pt}
\subsection{Evaluation} 
\begin{table}[!t]
\centering    

\resizebox{0.75\columnwidth}{!}{\begin{tabular}{l|c|c}
\toprule
Approach & $\downarrow$ PLR & $\uparrow$ ACC \\
\midrule
No Privacy Preservation & 100\% & 87.201\% \\  
\midrule
Gaussian Blur 1 & \textit{2.150\%} & 20.968\% \\
Gaussian Blur 2 & 88.978\% & \textit{77.778\%} \\ 
Trade-off Framework & 33.020\% & 71.827\% \\ 
Optical Flow &  1.030\%  & 16.308\%   \\
Image Swapping &  3.674\%  &  64.964\%  \\
\midrule
\rowcolor{lightgray}
Ours & 2.016\% & \textbf{78.843\%} \\
\bottomrule
\end{tabular}}
\caption{Privacy Leakage Ratio (PLR) and Utility Task Accuracy (ACC) of different privacy-preserving approaches and without privacy preservation.}
\label{main_result}
\vspace{-7pt}
\end{table}

The experimental results are evaluated using our proposed privacy leakage evaluation mechanism and a utility task evaluation metric: privacy leakage ratio from \vpl, indicating the ability of privacy-preservation, and video-based \fer accuracy from \fu, indicating the performance of our framework. 
As illustrated in Section~\ref{exp_setup}, the dataset contains 91 actor identities. The ideal ratio of the \vpl result should be close to $\frac{1}{91}$, or approximately 1.099\%. Similarly, the accuracy of the privacy-preserved video-based \fer task should ideally be close to the accuracy of the standard video-based \fer task.

\begin{table}[!t]
    \centering
    \resizebox{\columnwidth}{!}{
    \begin{tabular}{c|c c c c c c | c}
    \toprule
        Exp & Ang& Dis & Fea& Hap & Neu & Sad & $\uparrow$ Avg\\ \hline
        NPP & 91.7\% & 98.8\% & 80.3\% & 98.3\% & 84.9\% & 80.8\% & 87.2\%\\
        Ours &93.6\%&81.2\%&80.0\%&96.7\%&80.7\%&40.9\% & 78.8\%\\
    \bottomrule
    \end{tabular}
    }
    \caption{Recognition accuracy of \fer for videos on each facial expression after applying our proposed methods. NPP stands for no privacy preservation.}
    \label{expression_result}
    \vspace{-9pt}
\end{table}

Table~\ref{main_result} lists the privacy leakage ratio and utility task accuracy.
The ``No privacy preservation'' row represents the \fu-only task, illustrating the performance upper bound for \fu without any privacy preservation.
When the privacy leakage ratio of Gaussian Blur 1 is 2.150\% on the CREMA-D test set, which is close to but higher than our framework's privacy leakage ratio, the video-based \fer accuracy is 20.968\%. 
Conversely, When the utility task accuracy of Gaussian Blur 2 is 77.778\%, which is close to but lower than our framework's utility task accuracy, the privacy leakage ratio reaches 88.978\%. Table~\ref{expression_result} shows the recognition accuracy of each expression before and after privacy enhancement.
The results of the two experimental approaches demonstrate the difficulty in balancing privacy preservation and utility tasks using blurring techniques.

To benchmark our framework further, we used three state-of-the-art privacy-preserving approaches and implemented them on the CREMA-D dataset. 
The trade-off framework ensures relatively high recognition accuracy; however, the facial identity in privacy-preserved videos can still be classified with relatively high accuracy. 
The privacy-preserving effect of the optical flow and image swapping is a byproduct. 
The optical flow approach achieves the lowest privacy leakage ratio but also obtains the lowest \fer accuracy. In the image swapping experiment, an author's face replaces all faces in the test set. The results also support the conclusion from~\cite{lopez2022deepfakes} that swapping cannot adequately transfer the expression features to the swapped faces.
\vspace{-7pt}
\subsection{Threat Model}
\vspace{-5pt}
\begin{table}[!t]
\centering    
\resizebox{0.80\columnwidth}{!}{\begin{tabular}{l|c|c|c}
\toprule
Approach & $\downarrow$ SSIM & $\downarrow$ PSNR & $\downarrow$ PLR \\
\midrule
Gaussian Blur 1 & 0.94 & 31.14& 72.45\%  \\
Gaussian Blur 2 & 0.96 & 32.54 & 94.58\% \\ 
Trade-off Framework & 0.57 & 19.97 & 46.19\% \\ 
Optical Flow &  \textbf{0.40}  & \textbf{9.59} & 5.91\% \\
Image Swapping &  0.50  & 15.46&  41.91\%  \\
\midrule
\rowcolor{lightgray}
Ours & 0.41 & 10.18 & \textbf{5.65\%}  \\
\bottomrule
\end{tabular}}
\caption{Quantitative results of the threat model for our method and baselines on CREMA-D testing set.}
\label{threat_model_res}
\vspace{-12pt}
\end{table}

In our framework, we consider a white-box attacker who can access all details except for the controllers $C_{hpr}, C_{lpr}$ and $C_{fc}$ which are removed during deployment. The attacker can use the outputs from our training set alongside the original training set to train a recovery model. 
Following recent approaches~\cite{mi2024privacy,mi2023privacy,ji2022privacy}, we employ a full-scale U-Net~\cite{ronneberger2015u} as the recovery model. Using the CREMA-D training set, the outputs of our framework, as well as those from other baselines, serve as training inputs for the U-Net. We train six U-Nets, one for ours and the other five for baselines,  using $L_1$ reconstruction loss in conjunction with the original videos. 
For quantitative comparison, we use structural similarity (SSIM), peak signal-to-noise ratio (PSNR), and our proposed PLR for privacy preservation evaluation. The results are shown in Table~\ref{threat_model_res}. It shows that our method, together with optical flow, significantly outperforms the other five methods of privacy preservation against recovery attacks. However, the optical flow fails on the utility task, while our method demonstrates the highest performance on utility task, as shown in Table~\ref{main_result}.

\vspace{-5pt}
\subsection{Ablation Study} 
\label{ablation}
\vspace{-2pt}

\begin{table}[ht]
\centering

\begin{threeparttable}
\resizebox{0.37\textwidth}{!}{
\begin{tblr}{
  row{1} = {c},
  row{2} = {c},
  row{3} = {c},
  row{4} = {c},
  cell{1}{1} = {r=2}{},
  cell{1}{2} = {c=5}{},
  cell{1}{7} = {r=2}{},
  cell{1}{8} = {r=2}{},
  vline{2,7-8} = {-}{},
  vline{3-6} = {2-8}{},
  hline{1,9} = {-}{0.1em},
  hline{3-8} = {-}{},
  hline{2} = {2-6}{},
}
\begin{sideways}Task No. \end{sideways} & Components  &   &  & & & PLR      & ACC     \\
   & \begin{sideways}T\&IT.\textcolor{purple}{\tnote{a}}\end{sideways} & \begin{sideways}PE\textcolor{purple}{\tnote{b}}\end{sideways} & 
   \begin{sideways}IBC.\textcolor{purple}{\tnote{c}}\end{sideways} & \begin{sideways}FC.\textcolor{purple}{\tnote{d}}\end{sideways} & 
   \begin{sideways}FCC.\textcolor{purple}{\tnote{e}}\end{sideways} &        &          \\
   1  & \xmark & \cmark   &  \cmark   &  \cmark  &   \cmark  & 12.993\% & 73.028\%   \\
   2  & \xmark  & \cmark   & \xmark & \cmark    &  \cmark     & 32.975\% & 72.429\%  \\
   3  & \xmark     &  \cmark  &   \cmark  & \xmark & \xmark   & 12.903\% & 54.487\% \\
   4  & \cmark   &   \cmark  &  \cmark   &  \cmark & \xmark  & 2.106\%  & 61.187\% \\
   5  & \cmark   &   \cmark  &  \cmark   & \xmark & \xmark  & 2.061\%  & 48.656\% \\
   6  & \cmark   & \cmark  & \xmark    & \xmark    & \xmark   & 31.989\% & 68.831\% \\
\end{tblr}}
\begin{tablenotes}
\scriptsize
\item[\textcolor{purple}{a}] Transformation and Inverse Transformation;
\item[\textcolor{purple}{b}] Privacy Enhancers;
\item[\textcolor{purple}{c}] Identity Budget Controller;
\item[\textcolor{purple}{d}] Feature Compensator;
\item[\textcolor{purple}{e}] Feature Compensator Controller.
\end{tablenotes}
\end{threeparttable}
\vspace{-6pt}
\caption{Ablation Study Overview.}
\label{abs_all}
\end{table}


\begin{figure}[!t]
    \centering
    \includegraphics[width=0.9\linewidth]{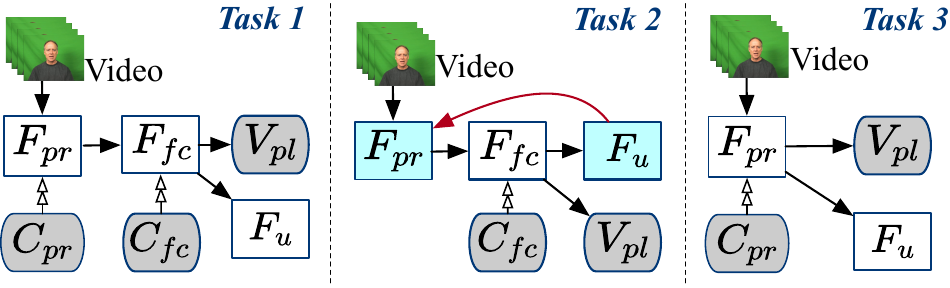}
    \caption{
    $F_{pr}$ and $C_{pr}$ denote the privacy enhancer and its controller. They work on the original video. The privacy enhancer and identity budget controller (in purple) share identical architectures and weights with \hpr and \chpr, respectively. In Task 2, the controller-free privacy enhancer is trained while updating \fu. 
    }
    \label{fig:ab1-3}
    \vspace{-10pt}
\end{figure}

\begin{figure}[ht]
    \centering
    \includegraphics[width=0.85\linewidth]{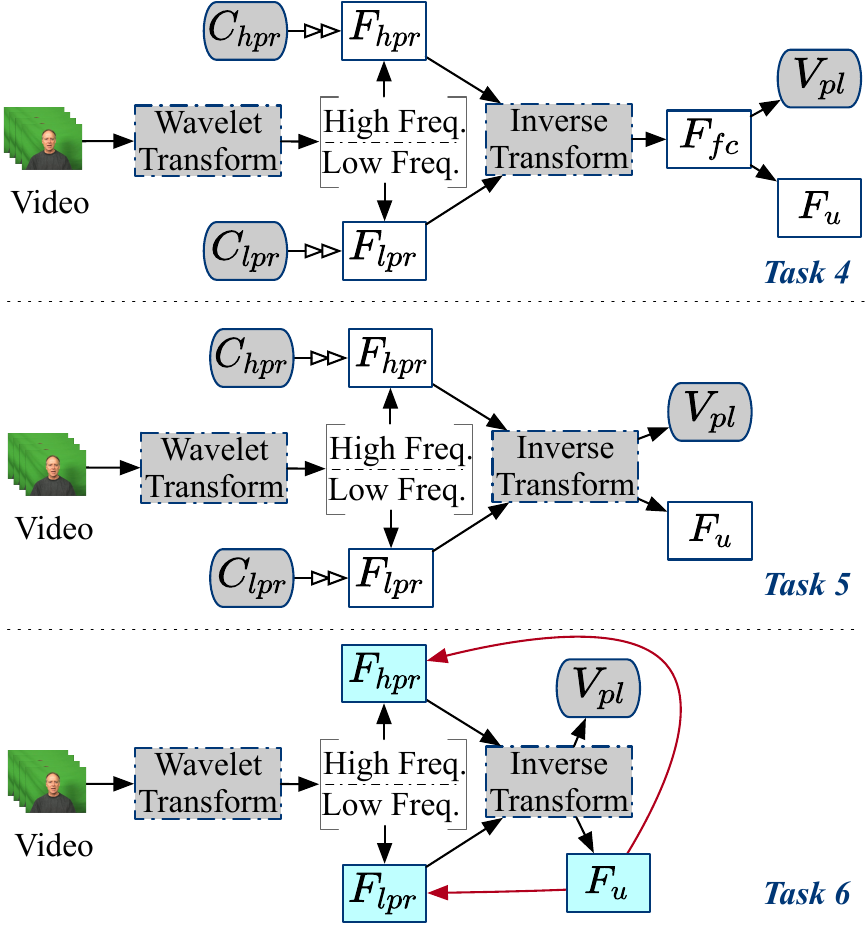}
    \caption{The structure of Tasks 4, 5 and 6. In Task 6, the parameters of controller-free \hpr and \lpr are updated while training \fu.}
    \label{fig:ab4-6}
\vspace{-10pt}
\end{figure}

Table~\ref{abs_all} shows the ablation study results, focusing on the components of controlled privacy preservation and controlled feature compensation, labeled as \Circled{1} and \Circled{2} in Figure~\ref{fig:framework}.
Figure~\ref{fig:ab1-3} and Figure~\ref{fig:ab4-6} illustrate the details of each ablation study task.

\textbf{Task 1} removes the wavelet transformation and inverse transformation steps in the controlled privacy preservation process. Without the separated privacy enhancer on high- and low-frequency components, the PLR increased. This demonstrates the importance of handling high- and low-frequency information separately to ensure effective privacy preservation.

\textbf{Task 2} further removes the identity budget controller,  
utilizing a bi-optimization approach in its absence. 
The results show a dramatic increase in privacy leakage, highlighting the crucial role of the identity budget controller in maintaining privacy while performing \fer.

\textbf{Task 3} removes the controlled feature compensation.
Unlike Task 1, it maintains the controlled privacy preservation using ResNet50 initialized with the same parameters as \chpr and \clpr, and a U-Net initialized with the same parameters as \hpr and \lpr. The absence of controlled feature compensation affects the balance between privacy preservation and utility task performance.

\textbf{Task 4} removes \cfc, and \fc is trained with mean squared error loss. The privacy leakage ratio remains at a similar level, with 46 identities correctly predicted, compared to 45 in our full framework. Without \cfc, the non-identity feature compensation is not as effective as required for the video-based \fer. 

\textbf{Task 5} removes both \cfc and \fc, eliminating the entire controlled feature compensation component. This corresponds to removing \Circled{2} from the framework in Figure~\ref{fig:framework}. The \fu is trained directly on the privacy-preserved videos from \Circled{1}. The video-based \fer performs sub-optimally on these privacy-preserved videos due to insufficient non-identity features.

\textbf{Task 6} further removes the identity budget controller from Task 5. \hpr, \lpr and \fu are trained as a whole. The transform, being a function with only hyperparameters, does not participate in the training process. The parameter update process is indicated by the red arrow. 
Despite the transformations applied, the absence of an identity budget controller results in sub-optimal performance for privacy enhancement.

The results of Tasks 1, 2, 3 and the full framework demonstrate that utilizing high- and low-frequencies privacy enhancement can effectively remove the identity information from the video. 
The results of Tasks 4 and 5 highlight the importance of the controlled feature compensation in improving the performance of video-based \fer. Our proposed controlled feature compensation effectively enriches the facial expression features, counteracting their potential loss during privacy enhancement.
Task 6 and our main results demonstrate that the privacy leakage validation mechanism can effectively evaluate privacy preservation approaches based on specific privacy attributes, such as the facial identity feature in our experiment. 

In addition, by observing the results of Tasks 2, 3, 5, and 6 for the privacy-preserving video-based \fer task, we conclude that our proposed transformation- and control-based privacy preservation approach can effectively remove specific privacy features. 
The results from bi-optimization approaches, specifically, Tasks 2 and 6 in Table~\ref{abs_all} and the Trade-off Framework in Table~\ref{main_result}, indicate that the bi-optimization approaches require further work to adapt to privacy-preserved video-based \fer.
Our framework decouples the privacy-preservation and utility tasks, connecting them via the controlled feature compensation, thereby enhancing the performance of both parts. 
From the results of Tasks 1 and 4 or 2 and 4, it is evident that features need to be enhanced under control rather than randomly.
Even after transformation, the facial identity and expression features cannot be completely separated, making it impossible to achieve the ideal utility task accuracy of 87.201\% as seen with no privacy preservation in Table~\ref{main_result}.

\vspace{-5pt}
\subsection{Limitation and Future work}
\vspace{-5pt}
Our framework has the following limitations. Firstly, the dataset must contain privacy labels, which are used to control the privacy enhancer during training. As a result, our framework cannot adapt to in-the-wild datasets without privacy labels. However, we have demonstrated the effectiveness and necessity of the controller for privacy preservation in Section~\ref{ablation}. In future research, we will further improve the framework to adapt to open-set scenarios and handle in-the-wild datasets.
Secondly, our privacy preservation focuses solely on facial identity attributes. There is still room to explore and adapt our framework for other utility tasks, such as pose estimation.
Thirdly, our utility task selects a basic video-analyzing neural network for video-based \fer. A state-of-the-art video \fer model could potentially further improve utility task accuracy. It is, however, not our main goal in this research.
A simple network, such as R3D used in this work, is sufficient to demonstrate the impact of our proposed privacy-preserving process on video-based FER tasks while remaining robust to the recovery attack. The small difference in accuracy between the privacy-preserving and non-privacy-preserving scenarios demonstrates the superiority of our framework.

\vspace{-1pt}

\vspace{-7pt}
\section{Conclusion}
\vspace{-3pt}
We propose a novel privacy-preserved video-based \fer framework that removes facial identity features from both low-frequency and high-frequency components via a wavelet transform, followed by an inverse transform. The framework compensates features in the inverse-transformed videos to generate feature-rich content for downstream utility tasks. Additionally, we introduce the privacy leakage ratio in the closed-set scenario, a mechanism for evaluating the privacy preservation approach. Our extensive experiments demonstrate that our framework achieves competitive results with a low privacy leakage ratio and high performance in video-based \fer, explaining the reasons for these results.

{\small
\bibliographystyle{ieee_fullname}
\bibliography{0.Ref}
}
\end{document}